\documentclass[sigconf]{acmart}

\AtBeginDocument{%
  }

\setcopyright{acmlicensed}
\copyrightyear{2024}
\acmYear{2024}
\acmDOI{XXXXXXX.XXXXXXX}

\acmConference[CIKM '24]{Proceedings of the 33nd ACM International Conference on Information and Knowledge Management}{October 21--25, 2024}{Boise, Idaho, USA} 

\usepackage{multirow}
\usepackage{array}
\usepackage{booktabs} 
\usepackage{multirow}
\usepackage{tabularx}
\usepackage{graphicx}
\usepackage{makecell}
\usepackage{caption}
\usepackage{subcaption}
\usepackage[titletoc]{appendix}
\usepackage{textcomp}
\usepackage{xspace}
\usepackage{soul} 
\usepackage{color}
\usepackage{xcolor}
\usepackage{bm}
\definecolor{shadecolor}{rgb}{0.92,0.92,0.92}  
\usepackage{tcolorbox} 
\usepackage[export]{adjustbox}
\definecolor{mygray}{RGB}{230,230,240}
\definecolor{myblue}{RGB}{175, 238, 235}
\definecolor{deepgreen}{rgb}{0.0, 0.5, 0.0}
\begin{document}

\title{UniMEL: A Unified Framework for Multimodal Entity Linking with Large Language Models}

\newcommand{\ours}{UniMEL\xspace}

\author{Qi Liu}
\authornote{Both authors contributed equally to this research.}
\affiliation{%
  \institution{ University of Science and Technology of China \& State Key Laboratory of Cognitive Intelligence}
  \city{Hefei}
  \country{China}
}
\email{liuqilq@mail.ustc.edu.cn}

\author{Yongyi He}
\affiliation{%
  \institution{ University of Science and Technology of China \& State Key Laboratory of Cognitive Intelligence}
  \city{Hefei}
  \country{China}
}
\email{vagabond@mail.ustc.edu.cn}
\authornotemark[1]

\author{Defu Lian}
\authornote{Corresponding Author.}
\affiliation{%
  \institution{ University of Science and Technology of China \& State Key Laboratory of Cognitive Intelligence}
  \city{Hefei}
  \country{China}
}
\email{liandefu@ustc.edu.cn}

\author{Zhi Zheng}
\affiliation{%
  \institution{ University of Science and Technology of China \& State Key Laboratory of Cognitive Intelligence}
  \city{Hefei}
  \country{China}
}
\email{zhengzhi97@mail.ustc.edu.cn}
\authornotemark[2]

\author{Tong Xu}
\affiliation{%
  \institution{ University of Science and Technology of China \& State Key Laboratory of Cognitive Intelligence}
  \city{Hefei}
  \country{China}
}
\email{tongxu@ustc.edu.cn}

\author{Che Liu}
\affiliation{%
  \institution{ University of Science and Technology of China \& State Key Laboratory of Cognitive Intelligence}
  \city{Hefei}
  \country{China}
}
\email{lc_1172@mail.ustc.edu.cn}

\author{Enhong Chen}
\affiliation{%
  \institution{ University of Science and Technology of China \& State Key Laboratory of Cognitive Intelligence}
  \city{Hefei}
  \country{China}
}
\email{cheneh@ustc.edu.cn}

\renewcommand{\shortauthors}{Qi Liu, Yongyi He et al.}

\begin{abstract}
Multimodal Entity Linking (MEL) is a crucial task that aims at linking ambiguous mentions within multimodal contexts to the referent entities in a multimodal knowledge base, such as Wikipedia. 
Existing methods focus heavily on using complex mechanisms and extensive model tuning methods to model the multimodal interaction on specific datasets. However, these methods overcomplicate the MEL task and overlook the visual semantic information, which makes them costly and hard to scale.
Moreover, these methods cannot solve the issues like textual ambiguity, redundancy, and noisy images, which severely degrade their performance. 
Fortunately, the advent of Large Language Models (LLMs) with robust capabilities in text understanding and reasoning, particularly Multimodal Large Language Models (MLLMs) that can process multimodal inputs, provides new insights into addressing this challenge. However, how to design a universally applicable LLMs-based MEL approach remains a pressing challenge.
To this end, we propose UniMEL, a \underline{uni}fied framework which establishes a new paradigm to process \underline{m}ultimodal \underline{e}ntity \underline{l}inking tasks using LLMs.
In this framework, we employ LLMs to augment the representation of mentions and entities individually by integrating textual and visual information and refining textual information. 
Subsequently, we employ the embedding-based method for retrieving and re-ranking candidate entities. 
Then, with only \textasciitilde0.26\% of the model parameters fine-tuned, LLMs can make the final selection from the candidate entities.
Extensive experiments on three public benchmark datasets demonstrate that our solution achieves state-of-the-art performance, and ablation studies verify the effectiveness of all modules.
Our code is available at \href{https://github.com/Javkonline/UniMEL}{https://github.com/Javkonline/UniMEL}.

\end{abstract}



\maketitle

\section{Introduction}

\begin{figure}[tp]
  \centering
  \includegraphics[width=\linewidth]{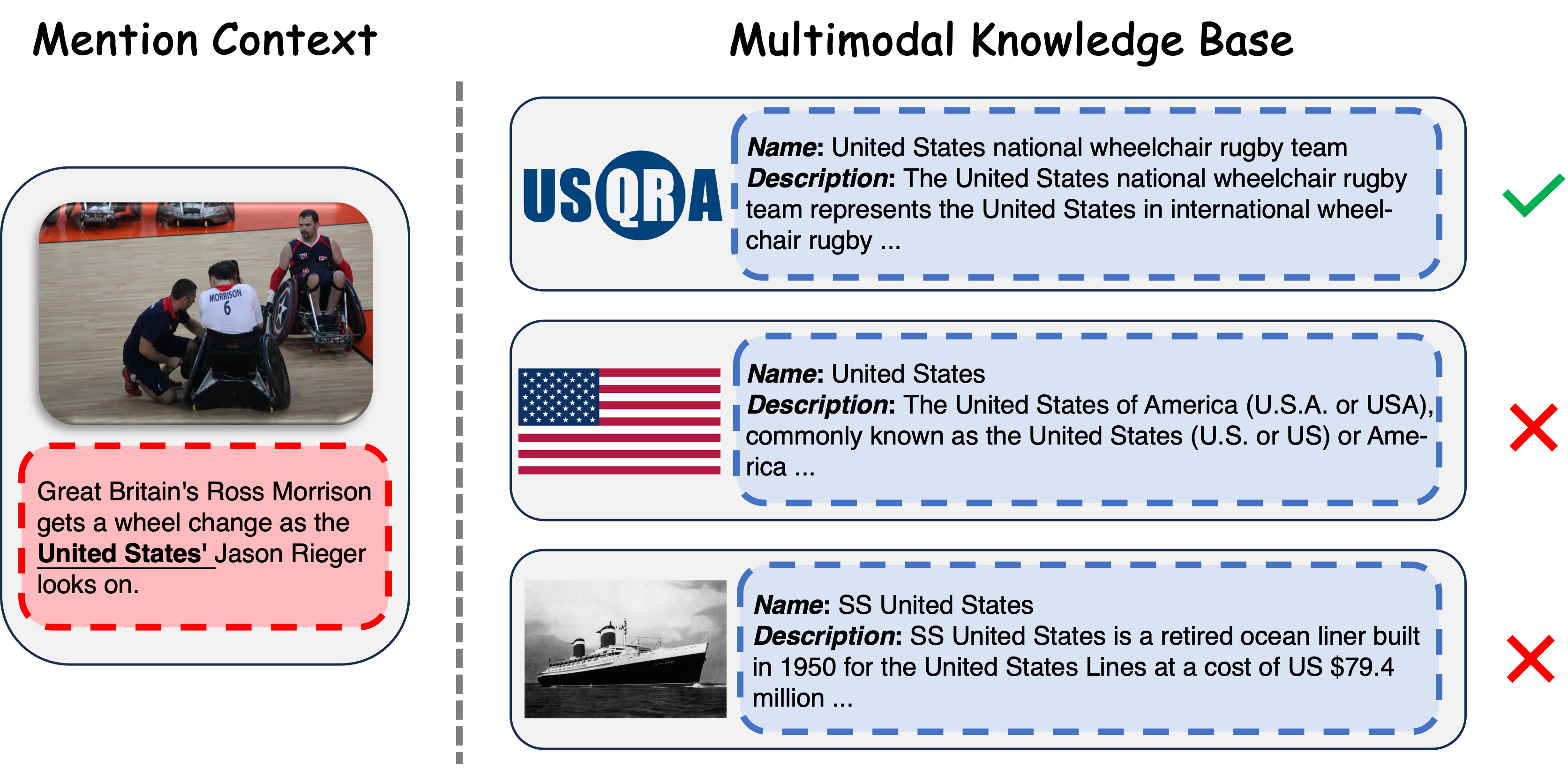}
  \caption{An example of Multimodal Entity Linking. Left: mention context, including mention image and mention text (highlighted in \textcolor[rgb]{1,0,0}{red}). Right: the similar candidate entities in Multimodal Knowledge Base, each with its entity image and entity text (highlighted in \textcolor[rgb]{0,0,1}{blue}).}
  \label{fig:example1}
\end{figure}

\textbf{Entity Linking} 
is a basic task in Knowledge Graph (KG) domains~\cite{chen2024knowledge}, which aims at linking mentions (i.e., segments of text referring to specific entities) of a document to entities in a knowledge base (KB). 
It is widely used in many Natural Language Processing (NLP) downstream applications, such as question answering~\cite{longpre2021entity, xiong2019improving, de-cao2019question}, recommendation systems~\cite{Geng2022Recommendation} and so on. 
Recently, the development and widespread use of social media and the Internet have further propelled textual and visual multimodality to become an important medium for data tasks. 
Meanwhile, the quality of online information is increasingly uneven, with many mentions being ambiguous, and their context being coarse. 
Therefore, in many cases, relying solely on textual modality to disambiguate the mentions proves inadequate.~\cite{moon2018multimodal, gan2021multimodal}. 
However, integrating textual and visual modality often facilitates a more precise and effortless disambiguation. 
Therefore, \textbf{Multimodal Entity Linking}, linking mentions along with textual and visual modality to a multimodal knowledge base (MMKB)~\cite{gan2021multimodal, Chen2024Building}, becomes essential. 
For example, as shown in Figure \ref{fig:example1}, the mention "\textit{United States'}" could be challenging to distinguish, as it may refer to various entities such as a country name, a sport team, or a liner. 
However, considering textual and visual information simultaneously, it becomes easier and clearer to accurately link the mention "\textit{United States'}" to the entity "\textit{United States national wheelchair rugby team}".

So far, many deep learning-based methods have been developed to fuse visual information with textual contexts when linking the multimodal mentions to entities. 
While these methods have made inspiring results in MEL tasks by fusing mention text and image to obtain mention representation~\cite{song2024dual}, applying cross-attention mechanisms~\cite{wang2022multimodal}, and encoding the image to extract features~\cite{song2024dual, shi2023generative}, these deep learning-based methods of MEL tasks still have several challenges as follows:
\begin{itemize}

\item\textbf{Redundant entity descriptions.} 
Generally, the description of the entity is usually too long, leading to a hard focus on valid information in the process of disambiguation. 
In this case, it is necessary to pay more attention to entities and mentions related parts.

\item\textbf{Lack of important semantic information in mentions.} 
Mentions differ from entities in that they lie in the contextual completeness of the feature information they contain. 
As shown in Figure \ref{fig:example1}, the mention textual context is a truncated piece extracted directly from documents, lacking pivotal semantic information and sufficient evidence for linking the mention to a specific entity effectively. 
How to utilize images from mentions to supplement their lacking semantic information becomes essential.

\end{itemize}
\vspace{0.5em}

Fortunately, Recent advances in Large language models (LLMs), such as GPT series models~\cite{radford2018improving,radford2019language,Brown2020Advances,achiam2023gpt} and LLaMA series models~\cite{radford2018improving,touvron2023llama,touvron2023llama2,llama3}, have considerably enhanced numerous NLP tasks~\cite{Raffel2019ExploringTL,qian2024unidm,Zhankui2023}. 
Pretrained on massive corpora, LLMs have the potential to generate instruction-following text via in-context learning~\cite{Brown2020Advances} or fine-tuning~\cite{ouyang2022training} in response to user prompts.
Although LLMs have demonstrated surprise on many data tasks, they inherently have a limited capacity to access visual information. 
Concurrently, Large Vision Models (LVMs) have the capability for comprehensive visual interactions, but commonly lag in reasoning.
In light of this complementarity, Multimodal Large Language Models (MLLMs) arise at an opportune time. 
The emergence of recent models such as GPT-4V~\cite{2023gpt4v}, LLaVA~\cite{Liu2023Advances}, Qwen-VL~\cite{bai2023qwen}, and text-to-image generation models like DALL-E-3~\cite{betker2023improving}, has marked significant progress in the MLLMs field. 
In addition, MLLMs have wide research scenarios, including image caption, vision question answering, etc. 

Based on the generative capabilities of LLMs, the previous methods let LLMs play different roles, such as an answer generator~\cite{shi2023generative} and memory controller~\cite{zhao2024ovel}, which has made inspiring results in the MEL task. 
However, utilizing LLMs for the MEL task is still untrivial due to the following reasons:

\vspace{0.5em}
\begin{itemize}

\item\textbf{Combining the visual context with the textual context effectively is challenging.} 
The images and textual content of the mentions usually correspond and complement each other, each providing additional information for mentions. 
Therefore, the fusion of images and text is quite important for entity linking.

\item\textbf{LLMs lack domain-specific knowledge.}
Although LLMs demonstrate powerful general capabilities, they do not directly excel in specific domain tasks (e.g., MEL tasks).

\end{itemize}
\vspace{0.5em}

To tackle these issues above, we introduce a unified framework for the MEL task with Large Language Models (\ours). 
Drawing inspiration from question-answering systems, mentions and entity candidates are concatenated to form queries inputted into the LLMs. 
Specifically, we aim to enhance the query, which consists of mentions and entity candidates, through the following ways:
(1) For mentions, the image and the contextual information associated with mentions are processed as input to MLLMs in order to extract deeper semantic relationships between the image and its context. 
This approach ensures that the integrity of the original image is maintained (i.e., without cropping or encoding), thereby fully leveraging the unaltered raw data. 
Considering the extensive corpus utilized for pre-training MLLMs, this method has the potential to enrich the concrete information pertaining to the mention.
(2) For entities, the overdetailed and redundant descriptions significantly challenge the MEL task. 
Short and precise new descriptions could be obtained through the utilization of LLMs' summarization capabilities.
(3) For narrowing down the candidate set, the embedding model is employed to retrieve and rerank the primitive candidate set. 
Subsequently, the top-K candidates are selected and concatenated with mentions to generate a multi-choice query. 
(4) For augmenting the capabilities of LLMs in MEL tasks, we finetune only a few parameters of the LLMs as a selector to choose the referent entity of the mention. 

Overall, the contributions of our work can be summarized as follows:
\begin{itemize}

\item In this work, we propose \ours, a framework to deal with the MEL task with LLMs and MLLMs, which could adequately fuse image and context of multimodal mentions and generate new concise descriptions of entities. 
To the best of our knowledge, it is the first work to introduce MLLMs-based methods in the MEL task.

\item We propose a unified prompt template set, specifically crafted for the Multimodal Entity Linking (MEL) task, to effectively bridge the semantic gap between modality content. 
And we finetune the LLM in the training set of the MEL dataset, to supply domain-specific knowledge.

\item We devise an ingenious solution to supplement the mention and refine the  entity information. 
Specifically, we provide the description of the mentions for the three publicly available datasets and present a summary of the descriptions of the entities. 

\item \ours achieves state-of-the-art (SOTA) results on three public multimodal entity linking datasets (22.3\% Top-1 accuracy gains on Richpedia, 21.3\% Top-1 accuracy gains on WikiMEL, 41.7\% Top-1 accuracy gains on Wikidiverse), exhibiting high performance.

\end{itemize}

\begin{figure*}[!h]
  \centering
  \includegraphics[width=0.95\linewidth]{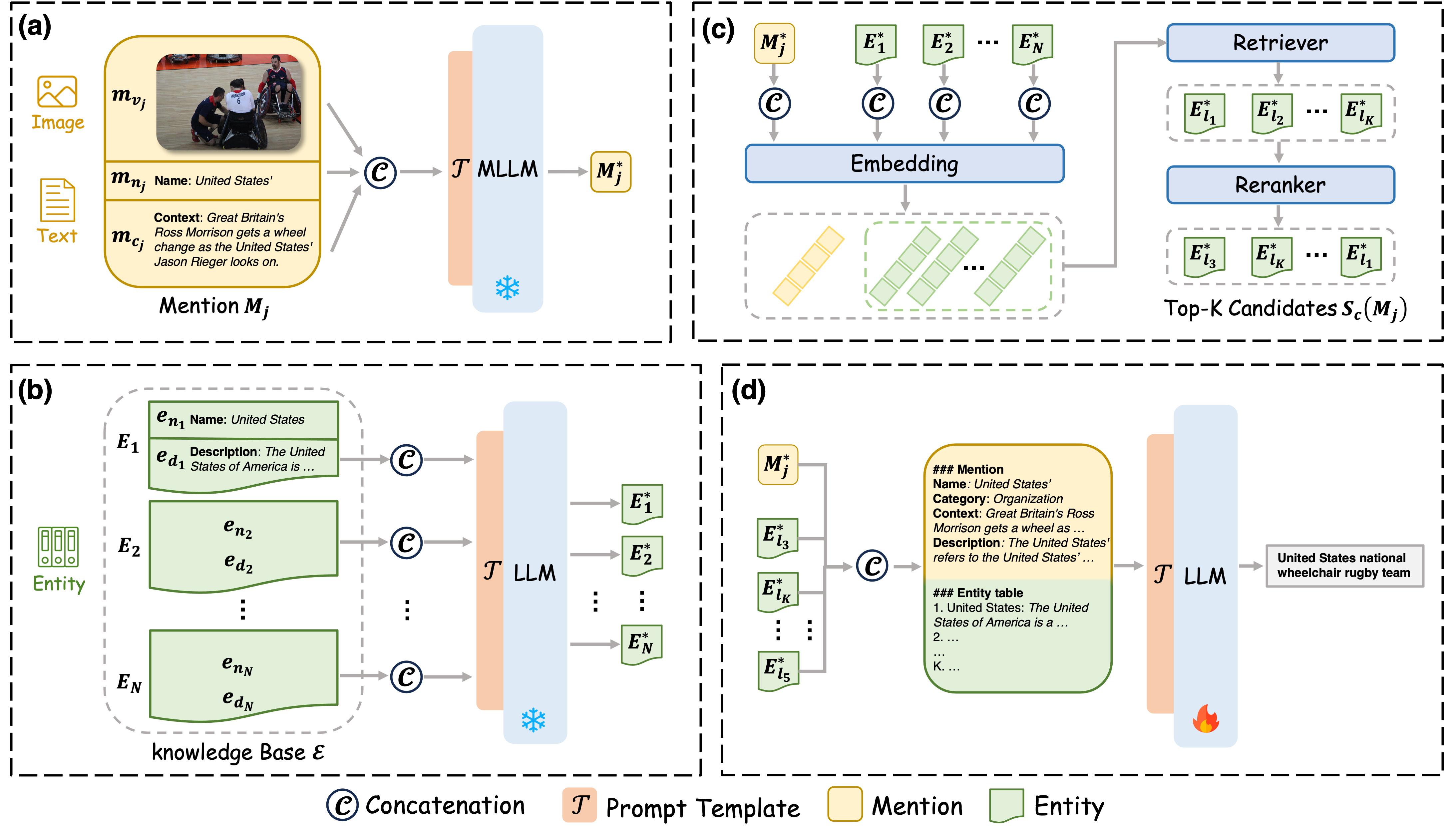}
  \vspace{-1.em}
  \caption{An overview of the \ours framework, which consists of four modules: (a) MLLMs-based Mention Augmentation, (b) LLMs-based Entity Augmentation, (c) Retrieval Augmentation and (d) Multi-choice Selection. Input consists of mention and entities, the frozen MLLM is applied to generate the mention description and the frozen LLM is applied to summary the entities description. And the tuned LLM is applied to select the referent entity for the mention.}
  \label{fig:overview2}
  \vspace{-1.em}
\end{figure*}

\section{Related work}

\subsection{Entity Linking}
Existing methods of entity linking can be divided into two steps: candidate selection and candidate reranking. For the first step, they commonly used coarse-grained filtering for generation candidates, e.g., TF-IDF~\cite{AIZAWA200345} and word2vec~\cite{Mikolov2013EfficientEO}. 
For the second step, they employed LSTM~\cite{hochreiter1997long} or BERT~\cite{kenton2019bert} for text encoders to measure the similarity, such as dot product~\cite{ganea2017deep} and cosine similarity~\cite{gillick2019learning} , and some research leveraged CNN~\cite{francis2016capturing}, cross-encoder and bi-and cross-encoder to capture the correlation between mention context and entity description, like BLINK~\cite{wu2020scalable}. 
By contrast, De Cao et al.~\cite{de2020autoregressive} used an autoregressive entity retrieval method by using BART~\cite{lewis2020bart} architecture to generate the unique name for each entity.
Although these methods have demonstrated significant effectiveness in text-only entity linking, accurately linking entities in short and coarse texts remains challenging. 
This has motivated the exploration of Multimodal Entity Linking.

\subsection{Multimodal Entity Linking}
To tackle the challenges above, MEL is essential. 
With the importance and abundance of multimodal information, it is vital to fuse visual and textual information during multimodal entity linking. 
How to reasonably fuse visual and textual information becomes serious. 
Moon et al.~\cite{moon2018multimodal} built a deep zero-shot multimodal network for MEL. 
Adjali et al.~\cite{adjali-etal-2020-building} proposed a novel method to build annotated datasets for evaluating methods on the MEL tasks and released a new MEL dataset, the Twitter MEL dataset. 
Zhou et al.~\cite{zhou2021weibo} released three MEL datasets and proposed a MEL dataset construction approach. 
Gan et al.~\cite{gan2021multimodal} modeled the alignment of textual and visual mentions as a bipartite graph matching problem. They also published a new MEL dataset, i.e., M3EL. 
Wang et al.~\cite{wang2022multimodal} extracted the hierarchical features of text and visual co-attention through the multimodal co-attention mechanism. 
To resolve the limited contextual topics and entity types, Wang et al.~\cite{wang2022wikidiverse} presented Wikidiverse, a high-quality human-annotated MEL dataset from Wikinews, including diversified contextual topics and entity types.
To model visual and textual information at the knowledge level, Zhang et al.~\cite{zhang2022multimodal} proposed an Interactive Learning Network to fully use the multimodal information.
Xing et al.~\cite{xing2023drin} exploited the fine-grained and dynamic alignment relations between entity and mention.
Luo et al.~\cite{luo2023multi} proposed a novel to understand the comprehensive expression of abbreviated textual context and implicit visual indication.
Shi et al.~\cite{shi2023generative} applied LLMs to the MEL task.
Song et al.~\cite{song2024dual} refined the queries with multimodal data.

\subsection{LLMs-based Data Tasks}
Large Language Models (LLMs) are pre-trained on the massive corpus, which has demonstrated outstanding capabilities in many data tasks, such as text summarization and domain-specific question answering. 
For summarization tasks, LLMs can transfer a long text to a brief text~\cite{singh2024indicgenbench}. 
Fine-tuning LLMs with domain-specific data enhances their performance on domain-specific tasks, such as ChatLaw~\cite{Cui2023ChatLawOL}, HuatuoGPT~\cite{zhang2023huatuogpt} and InstructGPT~\cite{ouyang2022training}. 
However, LLMs are susceptible to hallucination, often generating inaccurate and erroneous information.
To mitigate these problems, Lewis et al.~\cite{lewis2020retrieval} introduced the Retrieval-Augmented Generation (RAG) model, and then RAG enhances LLMs by retrieving relevant document chunks from the external knowledge base.
The development of multimodal large language models (MLLMs), which extend traditional LLMs to the multimodal domain, is increasingly gaining interest.
Similarly, MLLMs are pretrained on extensive vision and language data, which show promising ability in multimodal tasks, such as multimodal information summarization~\cite{cao2024characterizing}, image caption~\cite{li2023blip2, wang2022ofa, liu2023improvedllava, liu2023llava}, and visual question answering~\cite{dai2024instructblip}. 
Like LLMs, These MLLMs exhibit an inclination to generate hallucinations. Existing methods mitigate hallucination from four aspects: 
(1) Data, e.g., LRV-Instruction~\cite{liu2023mitigating}, 
(2) Model, e.g., LLaVA to LLaVA-1.6~\cite{liu2024llavanext}, 
(3) Training, e.g., RLHF-V~\cite{yu2023rlhf}, HA-DPO~\cite{zhao2023beyond} and 
(4) Inference, e.g., GCD~\cite{deng2024seeing}. 

\vspace{-0.5em}
\section{Preliminary}

\begin{table}[!t]
\caption{Compared with the description generated by MLLM and original source description. \textcolor{red}{Red} means the noise information for the MEL task, \textcolor{blue}{blue} is the valid information for the MEL task. }
  \label{tab:entity-no-image}
\begin{tabular}{>{\centering\arraybackslash}m{2cm}>{\centering\arraybackslash}m{5cm}}
\toprule 
\textbf{Attributes} &\textbf{Values} \\ 
\midrule 
\textbf{Image} & \includegraphics[width=2.5cm]{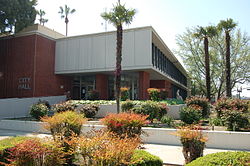}  \\
\midrule
\textbf{Entity} &  \centering\arraybackslash\scriptsize{Bakersfield City Hall}\\
\midrule
\textbf{Description (MLLM)} & \raggedright\arraybackslash\scriptsize{
The building is a large, \textcolor{red}{modern structure with a flat roof} and a facade that includes a mix of red brick and white concrete. It is situated in a \textcolor{blue}{landscaped area} with well-maintained gardens and palm trees, \textcolor{red}{suggesting a warm climate.} The architecture of the building is contemporary, with clean lines and a functional design.}\\
\midrule
\textbf{Description (Source)} & \raggedright\arraybackslash\scriptsize{Bakersfield City Hall (which is also referred to as City Hall South) is \textcolor{blue}{the center of government for the City of Bakersfield}, California. It houses \textcolor{blue}{the Mayor's office} and \textcolor{blue}{the City Council chambers}. It is \textcolor{blue}{located in the Civic Center}, Downtown. }\\
\bottomrule 
\end{tabular}
\end{table}

\textbf{Multimodal Large Language Model.}
Compared with text-only LLM, MLLM can process not only text inputs but also other modalities of input data (e.g., image), providing a richer and more comprehensive multimodal understanding and reasoning capability. 
\textbf{Multimodal Knowledge Base.}
A multimodal knowledge base (MMKB) is composed of an entity set. 
And each entity has different modal information. In the MMKB, the static attributes (i.e., occupation, name) of each entity are encapsulated within the text descriptions. 
Conversely, the images of entities in the MMKB tend to exhibit a broader range of dynamic attributes (i.e., clothes color) compared to textual descriptions. 
If dynamic attributes are employed as descriptors for textual entities, it commonly results in a misleading focus on these attributes when recognizing the entity. 
As presented in Table \ref{tab:entity-no-image}, we provide a specific example for an entity. 
Compared to the descriptions from raw text, the image descriptions do not produce more static attributes; instead, they generate additional noise information, where "\textcolor{red}{red} text" indicates noise and "\textcolor{blue}{blue} text" signifies valid attribute information. 
Incidentally, the pertinent information output by MLLMs is already presented in the textual description.
Generally, the entity descriptions in MMKB are very long and redundant, which is not conducive to representing the entity effectively. 
\textbf{Ambiguous Mentions.}
The mentions are the segments of text referring to specific entities, but their textual contexts are truncated pieces extracted directly from documents, which may lack pivotal semantic information, leading to ambiguities and insufficient evidence for linking the mention to a specific entity effectively. 
For example, the surname "\textit{Trump}", when mentioned in various contexts, can refer to different individuals. 
Secondly, the images of mentions may contain not only specific mentions but also other irrelevant and potentially distracting noise information.

\section{methodology}

We introduce our \ours, which is a \underline{un}ified framework to process \underline{m}ultimodal \underline{e}ntity \underline{l}inking tasks using LLMs. 
As shown in Figure \ref{fig:overview2}, \ours takes the multimodal mention context and entity information as input and gives the top-K candidate entity names. 
In the Section \ref{3.1}, we first present the problem formulation. Subsequently, we provide the details of our \ours, which has four modules: 
LLMs-based Entity Augmentation (Section \ref{3.2}), 
MLLMs-based Mention Augmentation (Section \ref{3.3}), 
Retrieval Augmentation (Section \ref{3.4}), 
Multi-choice Selection (Section \ref{3.5}). 

\subsection{Problem Formulation} \label{3.1}
Multimodal Entity Linking is aiming at mapping mentions with multimodal contexts to its corresponding entity in a knowledge base. 
Formally, we define an entity set of the multimodal knowledge base with $N$ entities as $\mathcal{E} = \{\bm{E}_i\}_{i=1}^{N}$. 
And each entity  $\bm{E}_i$ is represented by a triple $(\bm{e}_{n_i}, \bm{e}_{d_i}, \bm{e}_{v_i})$, including its entity name $\textbf{e}_{n_i}$, entity description $\textbf{e}_{d_i}$ and entity image $\bm{e}_{v_i}$. 
The mention set is defined as $\mathcal{M}=\{\bm{M}_j\}_{j=1}^{D}$, where each mention $\bm{M}_j$ is represented by a triple $(\bm{m}_{n_j}, \bm{m}_{c_j}, \bm{m}_{v_j})$, comprising the mention name $\bm{m}_{n_j}$, textual context $\bm{m}_{c_j}$, visual context $\bm{m}_{v_j}$. 
The referent entity of $\bm{M}_j$ in knowledge base $\mathcal{E}$ is predicted through:
\begin{equation} \label{Eq.2}
    e^*(\bm{M}_j) = \mathop{\arg\max}\limits_{\bm{E}_i \in \mathcal{E}} \text{sim}(\bm{M}_j, \bm{E}_i),
\end{equation}
where $\text{sim}(\text{·})$ is the similarity between the mention and the entity with multimodal information.

\subsection{LLMs-based Entity Augmentation} \label{3.2}

To address the problem of excessive length and redundant information in entity descriptions, we employ LLMs to summarize the descriptions effectively.
LLMs have been pretrained on a massive corpus that includes diverse world knowledge, and many works~\cite{pu2023summarization, zhang2024benchmarking} have demonstrated that they possess the capability of them to generate high-quality summaries for long texts.
First, we need to provide the LLM with the entity name and its original description.
Secondly, we design a specific instruction to emphasize the conciseness and content requirements of the summaries to be generated.
Then, following the given instruction, the LLM will generate a new, concise, high-quality, and information-rich summary of the entity description, thereby augmenting entity information.

As shown in Figure \ref{fig:overview2}b, for each entity $\bm{E}_i$, we provide LLMs with the entity name $\bm{e}_{n_i}$ and description $\bm{e}_{d_i}$, subsequently tasking them to generate a summarized description for the entity, represented as:
\begin{equation} \label{Eq.4}
    \bm{e}_{s_i} = \mathcal{LLM}(\mathcal{C} (\bm{e}_{n_i}; \bm{e}_{d_i})),
\end{equation}
where $\mathcal{C} (X;Y)$ denotes the function that concatenates $X$ and $Y$. 
We apply the designed and universal prompt template $T_{ea}$ to do this summarization.

\vspace{1.2em}\label{tem.1}
\noindent\fbox{\parbox{0.98\linewidth}{
\textbf{Prompt $T_{ea}$}:\\
\textsl{
Please summarize the main content of the given entity in
one sentence, including entity name and description.\\
\textbf{Entity name:} Asia-Pacific Economic Cooperation\\
\textbf{Entity description:} The Asia-Pacific Economic Cooperation (APEC;  AY-pek) is an inter-governmental forum for 21 member economies...\\
}
\textbf{Response:}\\
\textsl{
APEC is an inter-governmental forum of 21 member economies in the Pacific Rim that promotes free trade and economic cooperation...
}}}

\subsection{MLLMs-based Mention Augmentation} \label{3.3}

Due to the powerful visual comprehension and instruction following capabilities of MLLMs, we employ them to augment the descriptive information of mentions accompanied by images. 
We input the image containing the specific mention into the MLLM and provide it with the mention name and textual context. 
Firstly, we concatenate the mention name with its context to form a coherent semantic unit, ensuring that the connection is both meaningful and logically consistent.
Secondly, as MLLMs are adept at providing detailed and comprehensive descriptions of images, we design a specific task instruction to guide MLLMs to focus on the mention itself, preventing the generation of irrelevant environmental (e.g., \textit{"an outdoor setting with greenery"}) or detailed descriptions (e.g., \textit{"wearing a dark suit with a white shirt"}).
To let the MLLM produce well-formatted mention descriptions, we leverage its in-context learning capabilities, and construct a designed mention description as the demonstration example, specifying the output format for the MLLM.
Then, the MLLM utilizes visual and textual information to generate a description for the mention, thereby augmenting its information.

Specifically, as shown in Figure \ref{fig:overview2}a, for each mention $\bm{M}_j$, we provide the MLLM with three inputs: the mention image $\bm{m}_{v_j}$, the mention name $\bm{m}_{n_j}$, and the mention textual context $\bm{m}_{c_j}$. 
Then, we drive the MLLM to generate a high-quality textual description for $\bm{M}_j$, which can be represented as:
\vspace{-0.2em}
\begin{equation} \label{Eq.5}
    \bm{m}_{d_j} = \mathcal{MLLM}(\mathcal{C} (\bm{m}_{v_j}; \bm{m}_{n_j}; \bm{m}_{c_j})),
\end{equation}
by a designed prompt template $T_{ma}$.

\vspace{1.2em}
\noindent\fbox{\parbox{0.98\linewidth}{
\textbf{Prompt $T_{ma}$}:\\
\centerline{\includegraphics[width=0.3\linewidth]{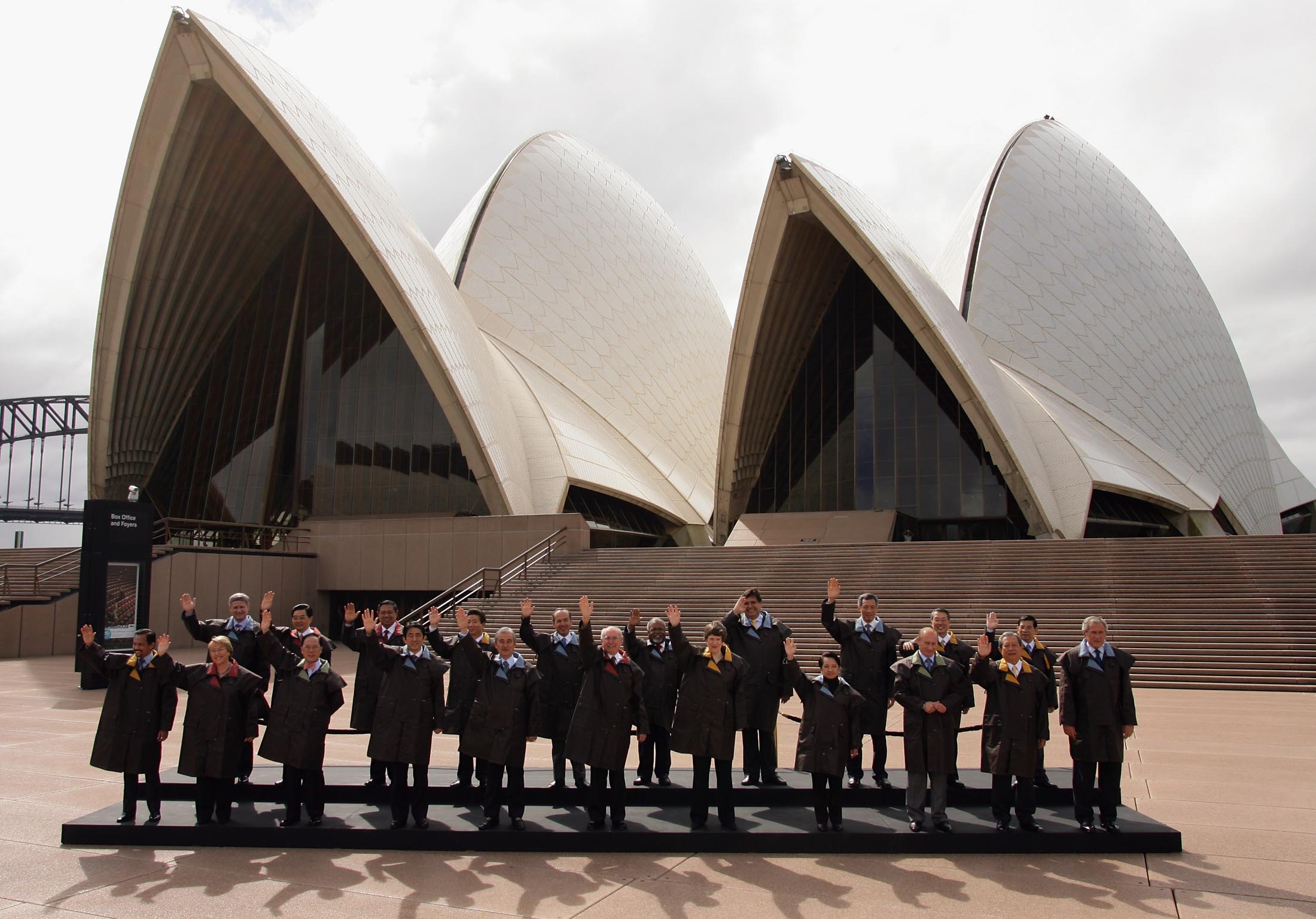}}\\
\textsl{
\textbf{Target entity name:} APEC.\\
\textbf{Image Description:} APEC Leaders wave for the media dressed in Driza-Bones in Sydney...\\
Only generate a description of the target entity, not a description of the image.\\
Answer follow the format: "The \textit{APEC} refer to..."\\
}
\textbf{Response:}\\
\textsl{
The APEC is a regional economic forum comprising 21 member economies in the Asia-Pacific region. It was established in 1989 to foster economic cooperation...
}
}}

\subsection{Retrieval Augmentation} \label{3.4}

Before selecting the entity that best matches the mention, it is generally necessary to narrow down the candidate set while ensuring its accuracy as much as possible. 
Thus, by focusing on a smaller, more precise set of candidates, the model can better learn the fine-grained distinctions between the mentions to be linked and different entities, thereby improving its performance.

With augmented information for entities and mentions, the first step is to concatenate the entity name $\bm{e}_{n_i}$ and new description $\bm{e}_{s_i}$, represented as:
\begin{equation} \label{Eq.6}
    \bm{e}_{r_i} = \mathcal{C}(\bm{e}_{n_i}; \bm{e}_{s_i}).
\end{equation}
Subsequently, we obtain the embedding representation of $\bm{e}_{r_i}$ using a pretrained embedding model, represented as:
\begin{equation} \label{Eq.7}
\bm{e}_{emb_i} = Embed(\bm{e}_{r_i}).
\end{equation}
By applying the above two steps to all entities in the knowledge base $\mathcal{E}$, we generate a vectorized knowledge base $\mathcal{V_E}$.

Then, for the mention $\bm{M}_j$, we concatenate the mention name $\bm{m}_{n_j}$ with its textual context $\bm{m}_{c_j}$ and description $\bm{m}_{d_j}$, represented as:
\begin{equation} \label{Eq.8}
    \bm{m}_{r_j} = \mathcal{C}(\bm{m}_{n_j}; \bm{m}_{c_j}; \bm{m}_{d_j}),
\end{equation}
Similarly, we derive the embedding representation of $\bm{m}_{r_j}$ using the pretrained embedding model, represented as:
\begin{equation} \label{Eq.9}
    \bm{m}_{emb_j} = Embed(\bm{m}_{r_j}).
\end{equation}
As shown in Figure \ref{fig:overview2}c, we compute the cosine similarity between the mention embedding $\bm{m}_{emb_j}$ and each entity embedding $\bm{e}_{emb_i}$ in a vectorized knowledge base $\mathcal{V_E}$, retrieving the $K$ entities with the highest similarity scores. 
These $K$ entities will subsequently serve the next module, which utilizes an LLM for entity selection. 
Some works~\cite{lu2022fantastically, zhao2021calibrate, hou2024large} indicate that in tasks involving multiple-choice selections by LLMs, the order in which optional objects are presented can influence the results. 
Thus, we rerank these K entities based on their similarity scores in descending order to obtain the set of candidate entities $\bm{S}_c(\bm{M}_j)$.

\subsection{Multi-choice Selection} \label{3.5}
LLMs exhibit strong general applicability, furthermore, fine-tuning them on domain-specific tasks and data could enhance their capabilities in handling domain-specific tasks.
Supported by the high-quality data and small-scaled candidates from the previous steps, we design a prompt template for the LLM instruction tuning. 
Formally, an LLM instruction can be formulated as a triplet: $(I,T,R)$, where $I$ signifies the instruction guiding the model, $T$ represents the text input provided, and $R$ denotes the ground-truth response serving as the target output during the fine-tuning process.
The LLM could predict an answer given the instruction and text input:
\begin{equation} \label{Eq.10}
    \mathcal{P} = \textit{f}(I,T|\theta),
\end{equation}
where $\mathcal{P}$ is the predicted answer of the LLM, and $\theta$ are the parameters of the model. In detail, we concatenate the mention $\bm{M}_j$ with the entities in its candidate set $\bm{S}_c(\bm{M}_j)$, obtaining the text input $T$:
\begin{equation} \label{Eq.11}
    \mathcal{C}(\bm{S}_c(\bm{M}_j)) = \mathcal{C}(\bm{E}_1; \bm{E}_1; ...; \bm{E}_k),
\end{equation}
\begin{equation} \label{Eq.12}
    \textit{T} = \mathcal{C}(\bm{M}_j;  \mathcal{C}\left(\bm{S}_c(\bm{M}_j\right))),
\end{equation}
where the $\mathcal{C}(\bm{S}_c(\bm{M}_j))$ denotes the concatenation of all entities in the candidate set. 
Additionally, we require the fine-tuned LLMs to possess a stable and manageable output format. 
We design an example of the desired format as a demonstration to better tune the formatting of LLMs outputs. 
We drive the LLM to select the entity that best matches the given mention by the following prompt template $T_{ms}$.

\vspace{1.2em}
\noindent\fbox{\parbox{0.98\linewidth}{
\textbf{Prompt $T_{ms}$}:\\
\textsl{
Your task is to create matches between the mention and entity tables to select the best-matched entity for the given mention.\\
\textbf{\#\#\# Mention}\\
\textbf{Name:} APEC\\
\textbf{Context:} APEC Leaders wave for the media dressed in Driza-Bones in Sydney...\\
\textbf{Description:} The APEC is a regional economic forum comprising 21 member economies in the Asia-Pacific region. It was established in 1989 to foster economic cooperation...\\
\textbf{\#\#\# Entity table}\\
1. Asia-Pacific Economic Cooperation: APEC is an inter-governmental forum of 21 members...\\
...\\
Output a JSON object following the format:\\
\textasciigrave\textasciigrave\textasciigrave json 
\{\{"id": "", "name": ""\}\}
\textasciigrave\textasciigrave\textasciigrave
}
\\
\textbf{Response:}\\
\textsl{
\textasciigrave\textasciigrave\textasciigrave json 
\{\{"id": "1", "name": "Asia-Pacific Economic Cooperation"\}\}
\textasciigrave\textasciigrave\textasciigrave
}
}}

\vspace{-0.5em}
\section{Experiments}

In this section, we conducted comprehensive experiments on three public MEL datasets to evaluate our proposed \ours. Moreover, we present extensive analyses to provide a more profound understanding of our framework.

\vspace{-0.5em}
\subsection{Experimental Setup}
\subsubsection{Datasets.} 
We begin with a review of datasets introduced in prior works. The first MEL dataset is SnapCaptionsKB, built by Moon et al.~\cite{moon2018multimodal}. 
It is composed of 12K user-generated image and textual caption pairs.
Adjali et al.~\cite{adjali-etal-2020-building} proposed a Twitter-MEL dataset, collected between January and April 2019 using Twitter official API.
Zhou et al.~\cite{zhou2021weibo} released three new MEL datasets: Weibo-MEL, Wikidata-MEL and Richpedia-MEL. Subsequently, Wang et al.~\cite{wang2022multimodal} constructed a new version of Twitter-MEL, WikiMEL and RichpediaMEL.
Gan et al.~\cite{gan2021multimodal} built the dataset M3EL from the Internet Movie Data Base (IMDb), The Movie Database (TMDb), and Wikipedia.
Later, Wang et al.~\cite{wang2022wikidiverse} released a high-quality MEL dataset WIKIDiverse.

Following previous works~\cite{song2024dual, shi2023generative, xing2023drin, luo2023multi, wang2023benchmarking}, we evaluated the performance of \ours on three MEL datasets: Wikidiverse, WikiMEL, and Richpedia~\cite{wang2020richpedia}:
\begin{itemize}

\item \textbf{Wikidiverse} 
is conducted from the Wikinews and covers different topics and 13 entity types (e.g., Person, Country, Organization, Event, Music, etc.), which is based on the KB of Wikipedia with about 16M entities in total.

\item \textbf{WikiMEL} 
contains over 22K multimodal samples extracted from Wikipedia and Wikidata. In contrast to Wikidiverse, the majority of entity types in WikiMEL are Person.

\item \textbf{Richpedia} 
collects the Wikidata index of entities in a large-scale multimodal knowledge graph Richpedia, and is obtained the multimodal information from Wikipedia.

\end{itemize}
We used Wikidata as our knowledge base and removed the mentions that we could not find the corresponding entities in Wikidata. 
We maintained the same dataset split consistent with previous work~\cite{song2024dual}. 
In Wikidiverse, 80\%, 10\%,10\% are divided into training set, validation set and test set. 
In WikiMEL and Richpedia, 70\%, 10\%, 20\% are divided into training set, validation set and test set. 
The statistics of the datasets are summarized in Table \ref{tab:exp4}.

\vspace{-0.5em}
\begin{table}[H]
\centering
  \caption{Statistics of three MEL datasets.}
  \label{tab:exp4}
\begin{tabular}{cccc}
\toprule
{\textbf{Statistic}}&Richpedia  & WikiMEL & Wikidiverse \\
\midrule
\# mentions & 18,752 &  25,846 &  16,097 \\
\# entities & 90,825 & 18,880& 72,288\\
\# images of mentions&15,852 &24,602 &12,743\\
\bottomrule
\end{tabular}
\end{table}
\vspace{-1em}

\begin{table*}[!t]
  \caption{Performance comparison of different methods on three MEL datasets from 100 candidate entities. The best score is highlighted in bold and the second best score is \underline{underlined}.}
  \centering
  \belowrulesep=0pt\aboverulesep=0pt
  \label{tab:exp}
\begin{tabular}{c|c|cccc|cccc|cccc}
\toprule
\specialrule{0em}{1pt}{0pt}
\multirow{2.4}{*}{\textbf{Modality}} & \multirow{2.4}{*}{\textbf{Model}} & \multicolumn{4}{|c|}{ \textbf{Richpedia} } & \multicolumn{4}{|c|}{ \textbf{WikiMEL} } & \multicolumn{4}{|c}{ \textbf{Wikidiverse} } \\[0.5mm]
\cmidrule(lr){3-6} \cmidrule(lr){7-10} \cmidrule(lr){11-14}

& & \multicolumn{1}{|c}{Top-1} & Top-5 & Top-10 & \multicolumn{1}{c|}{Top-20} & \multicolumn{1}{|c}{Top-1} & Top-5 & Top-10 & \multicolumn{1}{c|}{Top-20} & \multicolumn{1}{|c}{Top-1} & Top-5 & Top-10 & Top-20 \\ 
\specialrule{0em}{1pt}{0pt}
 \midrule
\specialrule{0em}{1pt}{0pt}
\multirow{4}{*}{Text-only} & BERT & 31.6 & 42.0 & 47.6 & 57.3 & 31.7 & 48.8 & 57.8 & 70.3 & 22.2 & 53.8 & 69.8 & 82.8 \\
& BLINK & 30.8 & 38.8 & 44.5 & 53.6 & 30.8 & 44.6 & 56.7 & 66.4 & - & 71.2 & - & - \\
& ARNN & 31.2 & 39.3 & 45.9 & 54.5 & 32.0 & 45.8 & 56.6 & 65.0 & 22.4 & 50.5 & 68.4 & 76.6 \\
\specialrule{0em}{1pt}{1pt}
\midrule
\specialrule{0em}{1pt}{1pt}
\multirow{10}{*}{Visual-text} & DZMNED & 29.5 & 41.6 & 45.8 & 55.2 & 30.9 & 50.7 & 56.9 & 65.1 & - & 39.1 & - & - \\
& JMEL & 29.6 & 42.3 & 46.6 & 54.1 & 31.3 & 49.4 & 57.9 & 64.8 & 21.9 & 54.5 & 69.9 & 76.3 \\
& MEL-HI & 34.9 & 43.1 & 50.6 & 58.4 & 38.7 & 55.1 & 65.2 & 75.7 & 27.1 & 60.7 & 78.7 & 89.2 \\
& HieCoAtt & 37.2 & 46.8 & 54.2 & 62.4 & 40.5 & 57.6 & 69.6 & 78.6 & 28.4 & 63.5 & 84.0 & 92.6 \\
& GHMFC & 38.7 & 50.9 & 58.5 & 66.7 & 43.6 & 64.0 & 74.4 & 85.8 & - & - & - & - \\
& MMEL & - & - & - & - & 71.5 & 91.7 & 96.3 & 98.0 & - & - & - & - \\
& CLIP & 60.4 & 96.1 & 98.3 & 99.2 & 36.1 & 81.3 & 92.8 & 98.3 & 42.4 & 80.5 & 91.7 & 96.6 \\
& DRIN & - & - & - & - & 65.5 & 91.3 & 95.8 & 97.7 & - & - & - & - \\
& DWE & 67.6 & 97.1 & \underline{98.6} & \underline{99.5} & 44.7 & 65.9 & 80.8 & 93.2 & 47.5 & 81.3 & 92.0 & 96.9 \\
& DWE$+$ & \underline{72.5} & \underline{97.3} & \bf{98.8} & \bf{99.6} & \underline{72.8} & \bf{97.5} & \bf{98.9} & \bf{99.7} & \underline{51.2} & \underline{91.0} & \underline{96.3} & \underline{98.9} \\
\specialrule{0em}{1pt}{1pt}
\midrule
\specialrule{0em}{1pt}{1pt}
Visual-text & \ours (ours) & \bf{94.8} & \bf{97.9} & 98.3 & 98.8 & \bf{94.1}  & \underline{97.2} & \underline{98.4} & \underline{98.9} & \bf{92.9} & \bf{97.0} & \bf{99.5} & \bf{99.8} \\
\specialrule{0em}{1pt}{1pt}
\bottomrule
\end{tabular}
\end{table*}

\subsubsection{Baselines.} 
We compared our method with recent state-of-the-art methods, which are divided into two groups: 
(1) text-only methods and 
(2) visual-text fusion methods. 
Specifically, we considered:

\begin{itemize}

\item \textbf{ARNN}~\cite{eshel2017named} (text-only) captures textual features by a duo of Attention-RNN.

\item \textbf{BERT}~\cite{devlin2018bert} (text-only) is used to encode the mention context and entity description and then calculate the relevance score.

\item \textbf{BLINK}~\cite{wu2020scalable} (text-only) integrates bi-encoder for candidate generation and cross-encoder for final entity disambiguation.

\item \textbf{HieCoAtt}~\cite{Lu2016HieCoAtt} (visual-text) proposes a co-attention mechanism and and constructs co-attention maps at different levels.

\item \textbf{DZMNED}~\cite{moon2018multimodal} (visual-text) takes a multimodal attention mechanism to fuse visual, textual and character-level
features of mention.

\item \textbf{JMEL}~\cite{adjali2020ecir} (visual-text) uses fully connected layers to project the visual and textual features into an implicit joint space.

\item \textbf{MEL-HI}~\cite{zhang2021attention} (visual-text) designs a two-stage image and text correlation mechanism to eliminate irrelevant images, and introduces an attention mechanism to capture feature in the mention and entity.

\item \textbf{CLIP}~\cite{Alec2021CLIP} (visual-text) employs two Transformer-based encoders to capture the relationships and semantics between visual and textual information.

\item \textbf{GHMFC}~\cite{wang2022multimodal} (visual-text) uses gated hierarchical multimodal fusion and contrastive training to learn cross-modality information at a fine-grained level.

\item \textbf{MMEL}~\cite{Yang2023MMEL} (visual-text) proposes a joint learning framework to learn the features of contexts and entity candidates together.

\item \textbf{DRIN}~\cite{xing2023drin} (visual-text) models four different types of alignment between a mention and entity and builds a dynamic GCN to select the corresponding alignment relations for different input samples.

\item \textbf{DWE}~\cite{song2024dual} (visual-text) views each mention as a query and enhances query by refined multimodal information. in addition, it enriches the semantics of entity representation by Wikipedia.

\end{itemize}

\subsubsection{Evaluation Metric.}
Following previous works~\cite{xing2023drin,song2024dual}, we use the Top-k accuracy metric for evaluation:
\begin{equation}
    \text{Top-}k=\frac{1}{D} \sum_{\bm{M}_j \in \mathcal{M}}\left[\sum_{\bm{E}_i \in \bm{S}_c(\bm{M}_j)}\text{I}\{\text{S}(\bm{M}_j, \bm{E}^{*})<\text{S}(\bm{M}_j, \bm{E}_i)\}<k\right],
\end{equation}
where $\mathcal{M}$ represents the set of mentions in the dataset, $\bm{S}_c(\bm{M}_j)$ denotes the set of candidate entities to be linked with mention $\bm{M}_j$, and $\bm{E}^{*}$ is the ground truth entity which should b linked with $\bm{M}_j$. $\text{I}$ is the indicator function, and $\text{S}$ is the similarity computation function.

\subsubsection{Candidate Retrieval.}
For a fair comparison, we employed the same method as recent works~\cite{wang2022multimodal, xing2023drin, song2024dual} for selecting the Top-100 candidate entities for each mention in a coarse-grained manner.
Specifically, in Richpedia and Wikimel, we calculated the similarity between entity names with mention names by fuzzy string matching\footnote{https://github.com/seatgeek/thefuzz} to select 100 candidate entities. 
In Wikidiverse, since the dataset provides 10 similar candidate entities, we only need to supplement additional 90 candidate entities using fuzzy string matching in the same way.

\subsubsection{Implementation Details.}
Our \ours framework is implemented with PyTorch on NVIDIA RTX A6000. 
We employed LLaMA-3-8B~\cite{llama3} and LLaVA-1.6~\cite{liu2024llavanext} as
our default LLM and MLLM, with the temperature set to 0 and other parameters remaining at their default settings, unless otherwise stated. 
For the embedding model, we employed SFR-Embedding-Mistral~\cite{SFRAIResearch2024}. 
We adopted cosine similarity to calculate the relevance between the embeddings of the mention information and the entity information. 
During the model fine-tuning, we sampled 10,000 mentions from the Wikidiverse training set and constructed the fine-tuning dataset according to our framework procedure. 
Subsequently, we fine-tuned LLaMA-3-8B using the LoRA~\cite{hu2022lora} method, setting the rank to 8 and alpha to 32. 
We leveraged the AdamW optimizer~\cite{loshchilov2018decoupled} with a batch size of 1, a learning rate of 1e-4 and a warmup ratio of 0.03.

\vspace{-0.5em}
\subsection{Main Results}

Table \ref{tab:exp} presents the performances of our proposed \ours method against several competitive approaches on Wikidiverse, WikiMEL and Richpedia datasets. 
We report Top-1, Top-5, Top-10 and Top-20 accuracy for all three datasets. 
Overall, based on these results, we can draw some observations and conclusions as follows:

First, our \ours achieves the best performance compared to existing SOTA methods on three datasets, suggesting its effectiveness.
Text-only methods, such as BERT, demonstrated impressive performance, and could even compete with some visual-text methods like DZMNED and JMEL.
It indicates the enduring importance of textual information as a fundamental modality for MEL, and the effective utilization of this modality is crucial in MEL tasks. 
It is observed that BLINK marginally underperforms compared to BERT on WikiMEL and Richpedia but performs better on Wikidiverse. 
They both extract the representations of mentions and entities through text encoders.
ARNN also achieves comparable results comparable to those of BERT by capturing text features with Attention-RNN.
Moreover, compared with the SOTA visual-text methods, the text-only methods still have a gap in performance. 
It is difficult to deal with vague and low-quality mention information using only textual information.

Furthermore, each visual-text method is noteworthy, showing diverse results. 
Their differences lie in the methods of extracting visual representations and integrating them with textual information. 
Among the first five visual-text methods shown in Table \ref{tab:exp}, GHMFC appears more competitive. 
This could be attributed to its fine-grained learning of cross-modal information. 
It suggests that shallow modality interactions and simplistic multimodal fusion may not enhance the performance of the MEL task. 
Subsequently, CLIP, pre-trained on a large-scale image-text corpus, yields notable results.
It is worth noting that DWE$+$ outperforms all other baselines on three datasets, coming second only to our proposed \ours. 
It benefits from its extensive utilization of visual information, complemented by the employment of Wikipedia for semantic augmentation in entity representation.
Overall, on Richpedia, WikiMEL and Wikidiverse, our \ours obtains the Top-1 accuracy of 94.8\%, 94.1\% and 92.9\%, respectively, improving 22.3\%, 21.3\% and 41.7\% over the previous methods.

\vspace{-0.2em}

\begin{table}[!h]
\centering
  \caption{Top-1 accuracy for the model from 16 candidates.}
  \vspace{-0.2em}
  \label{tab:exp2}
\begin{tabular}{ccc}
\toprule
\multirow{2.4}{*}{\textbf{Model}} & \multicolumn{2}{c}{\textbf{Top-1 Accuracy (\%)}}\\
\cmidrule(lr){2-3}
 & WikiMEL & Wikidiverse \\
\midrule
GEMEL & 82.6 & 86.3 \\
\ours &\bf{94.2} & \bf{93.0}\\
\bottomrule
\end{tabular}
\end{table}
\vspace{-1.5em}

\begin{table}[!h]
  \caption{Top-K accuracy for the model from 10 candidates.}
  \vspace{-0.2em}
  \label{tab:exp3}
\begin{tabular}{cccc}
\toprule
\multirow{2.4}{*}{\textbf{Model}} & \multicolumn{3}{c}{\textbf{Wikidiverse}} \\
\cmidrule(lr){2-4}
& Top-1 & Top-3 & Top-5 \\
\midrule
DRIN & 51.1 & 77.9 & 89.3 \\
\ours& \bf{93.8} & \bf{94.1} & \bf{97.7}\\
\bottomrule
\end{tabular}
\end{table}

\vspace{-0.2em}

In addition, we compared our \ours with the existing SOTA LLMs-based method, GEMEL~\cite{shi2023generative}, on Top-1 accuracy. 
GEMEL is the first work to introduce generative methods based on LLMs in the MEL task.
On WikiMEL and Wikidiverse, we aligned with GEMEL by setting the number of candidate entities to 16. 
As shown in Table \ref{tab:exp2}, our \ours continues to demonstrate superiority, outperforming GEMEL by 11.6\% on WikiMEL and 6.7\% on Wikidiverse. 
Meanwhile, we compared our \ours with the GCN-based method, DRIN, on Wikidiverse. 
Following the DRIN setup, we expressly set the number of candidate entities on Wikidiverse to 10 and tested for Top-1, Top-3, and Top-5 accuracy. 
As shown in Table \ref{tab:exp3}, it is evident that \ours outperforms DRIN by 42.7\% on Top-1 accuracy.
It can be observed that the smaller the number of candidate entities, the easier it is to select the link entity.

\begin{table}[h]
  \caption{Top-1 accuracy for different embedding models.}
  \vspace{-0.2em}
  \label{tab:embed}
\begin{tabular}{lcccccc}
\toprule
\multirow{2.4}{*}{\textbf{Model}} & \multicolumn{3}{c}{\textbf{Top-1 Accuracy (\%)}}\\
\cmidrule(lr){2-4}
 & Richpedia & WikiMEL & Wikidiverse\\
\midrule
SFR-Embedding-Mistral & 94.8 & 94.1 & 92.9 \\
GTE-large   & 94.7 & 94.3 & 88.9\\
\bottomrule
\end{tabular}
\end{table}

For fairness, we conducted tests on embedding models of various sizes. 
In addition to the SFR-Embedding-Mistral with 7B parameters, we also utilized the GTE-large~\cite{li2023towards} with 0.3B parameters for comparison. 
Table \ref{tab:embed} presents the Top-1 accuracy of these two models on Wikidiverse, WikiMEL, and Richpedia. 
Despite the significant reduction in the size of the embedding model, the Top-1 accuracy only slightly decreases on Wikidiverse but remains virtually unchanged on WikiMEL and Richpedia. 
This minor decrease on Wikidiverse can be attributed to its more complex entity types.
Since the augmented entity information possesses rich and high-quality content with similar semantic logic, there is less demand for a large-size embedding model.
It demonstrates the effectiveness of the Knowledge Augmentation and the generality of the Retrieval Augmentation in our \ours framework.

\vspace{-0.5em}

\begin{table}[h]
  \caption{Top-1 accuracy for different LLMs.}
  \label{tab:exp-LLMs}
\begin{tabular}{lccc}
\toprule
\multirow{2.4}{*}{\textbf{LLM}} & \multicolumn{3}{c}{\textbf{Top-1 Accuracy (\%)}}\\
\cmidrule(lr){2-4}
 & Richpedia & WikiMEL & Wikidiverse\\
\midrule
LLaMA3-8B (fine-tuned) & 94.3 & 93.8 & 93.5 \\
LLaMA3-8B  & 85.7 & 90.5 & 75.9 \\
Mistral-7B & 84.5 & 88.5 & 73.6\\
Qwen-1.5-7B & 84.9 & 86.4 & 71.2\\
gpt-3.5-turbo-0125 & 86.3 & 91.3 & 74.3 \\
gpt-4-turbo-2024-04-09 & 91.6 & 93.8 & 78.8\\
\bottomrule
\end{tabular}
\end{table}

To further analyse the generality of our \ours across different LLMs, we conducted experiments on various LLMs including the unfine-tuned LLaMA3-8B, Mistral-7B~\cite{jiang2023mistral}, Qwen-1.5-7B~\cite{qwen}, GPT-3.5 Turbo (gpt-3.5-turbo-0125), and GPT-4 Turbo (gpt-4-turbo-2024-04-09). 
As shown in Table \ref{tab:exp-LLMs}, without any fine-tuning, LLaMA3-8B performs comparably to GPT-3.5 Turbo, while GPT-4 Turbo achieves better results due to its larger model size.
The performance of Mistral-7B and Qwen-1.5-7B is inferior to that of LLaMA3-8B.
Moreover, with only \textasciitilde0.26\% of the model parameters fine-tuned, LLaMA3-8B achieves remarkable performance improvements, with increases of 8.6\%, 3.3\%, and 17.6\% on three datasets, respectively. 
It is evident that our \ours exhibits high generality and strong scalability, proving generally effective for the widely-used LLMs.

\subsection{Ablation Study}

\begin{table}[H]
  \caption{Performance of ablation studies on main modules and visual modality of UniMEL.}
  \label{tab:exp-ablation}
\resizebox{\linewidth}{!}{
\begin{tabular}{lccc}
\toprule
\multirow{2.4}{*}{\textbf{Model}} & \multicolumn{3}{c}{\textbf{Top-1 Accuracy (\%)}}\\
\cmidrule(lr){2-4}
 & Richpedia & WikiMEL & Wikidiverse\\
\midrule
\ours & 94.8 & 94.1 & 92.9 \\

w/o Multi-choice Selection & 94.2 & 89.8 & 71.2\\
w/o Retrieval Augmentation & 80.4 & 90.8 & 68.6\\
w/o Entity Augmentation & 93.5 & 91.9 & 90.6\\
w/o Mention Augmentation & 88.5 & 91.7 & 86.4\\
w/o E\&M Augmentation & 87.4 & 90.3 &85.8 \\
\midrule
w/o Visual Information & 88.6 & 92.4 & 90.1 \\
w/o Visual \& Selection & 80.9 & 88.1 & 69.7\\
\bottomrule
\end{tabular}
}
\end{table}

\begin{table*}[!t]
\caption{Case Study. For \underline{\textbf{underlined}} mentions, \textcolor{deepgreen}{green} and \textcolor{red}{red} predictions indicate correct and incorrect prediction results, respectively. The text-only prediction results are generated by \ours without using visual information.}

\vspace{-0.2em}

\belowrulesep=0pt\aboverulesep=0pt
  \label{tab:case-study}
\begin{tabular}{>{\centering\arraybackslash}m{2cm}>{\centering\arraybackslash}m{3cm}>{\centering\arraybackslash}m{3cm}>{\centering\arraybackslash}m{3cm}>{\centering\arraybackslash}m{3cm}}
\toprule 
\textbf{Case} &\textbf{1} &\textbf{2} &\textbf{3} &\textbf{4}  \\ 
\midrule 
\multirow{2.3}{*}{\textbf{Image}} & \adjustbox{padding=0cm 0.1cm}{\includegraphics[height=1.7cm,valign=c]{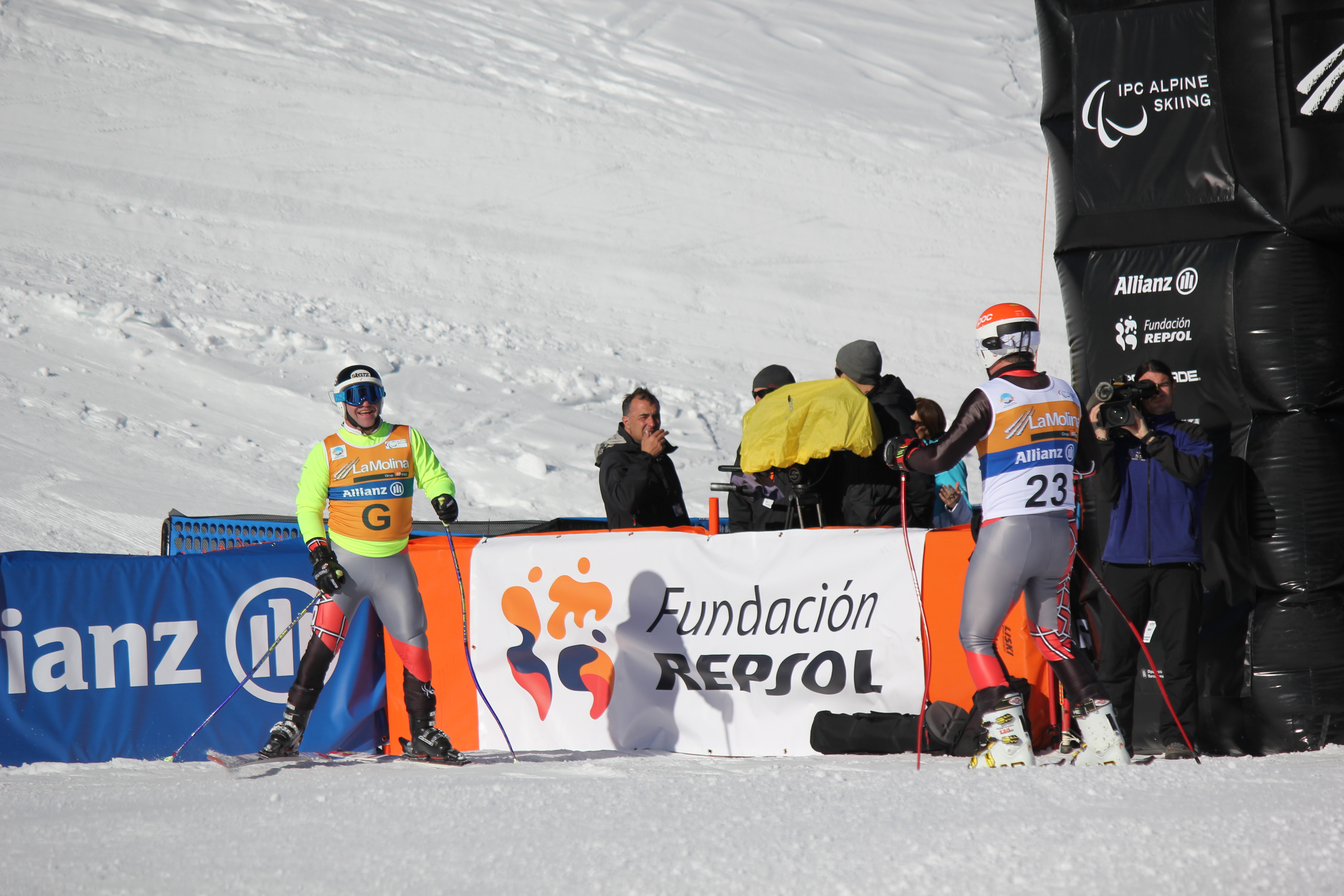}} & \adjustbox{padding=0cm 0.1cm}{\includegraphics[height=1.7cm,valign=c]{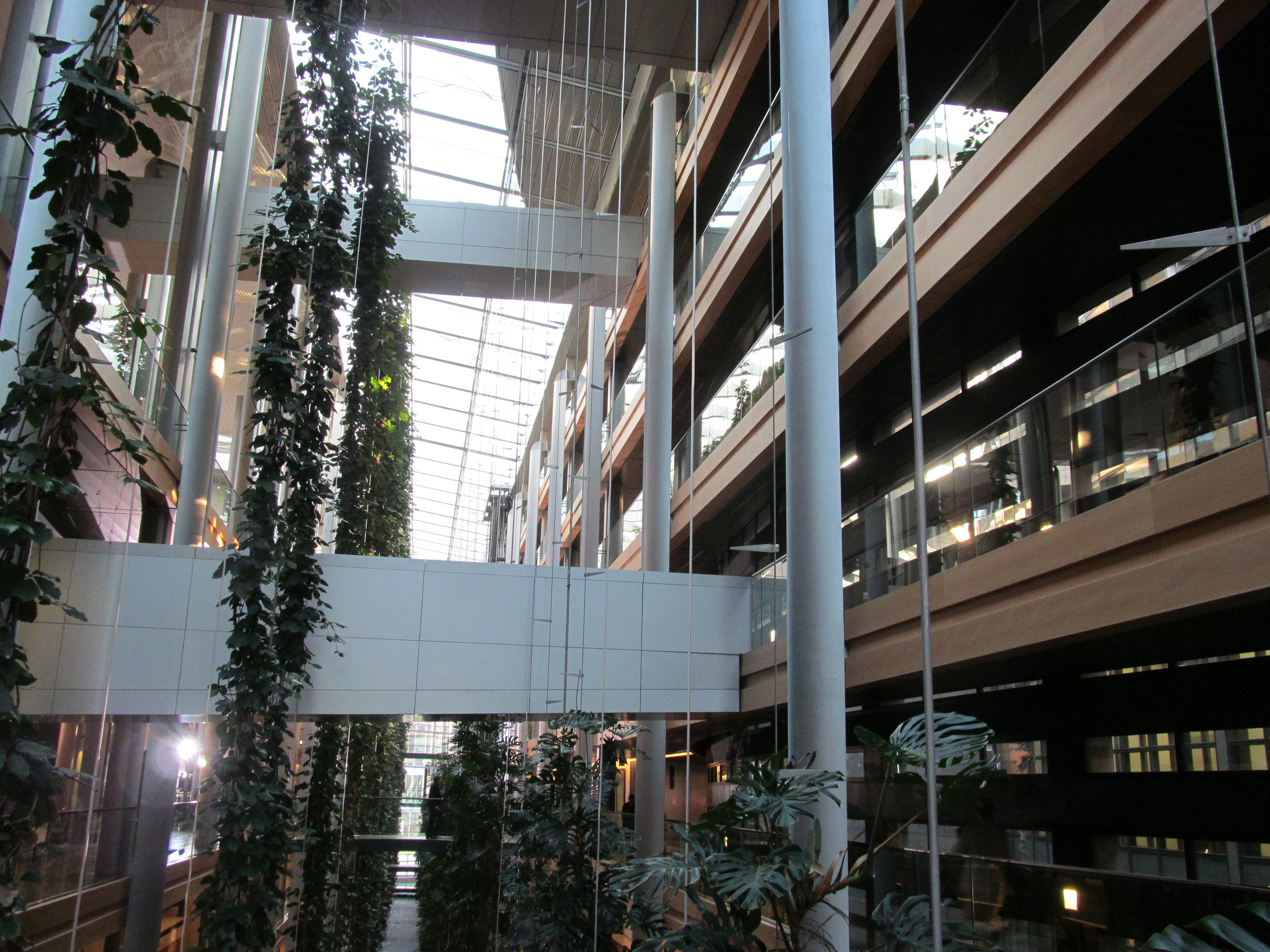}} & \adjustbox{padding=0cm 0.1cm}{\includegraphics[height=1.7cm,valign=c]{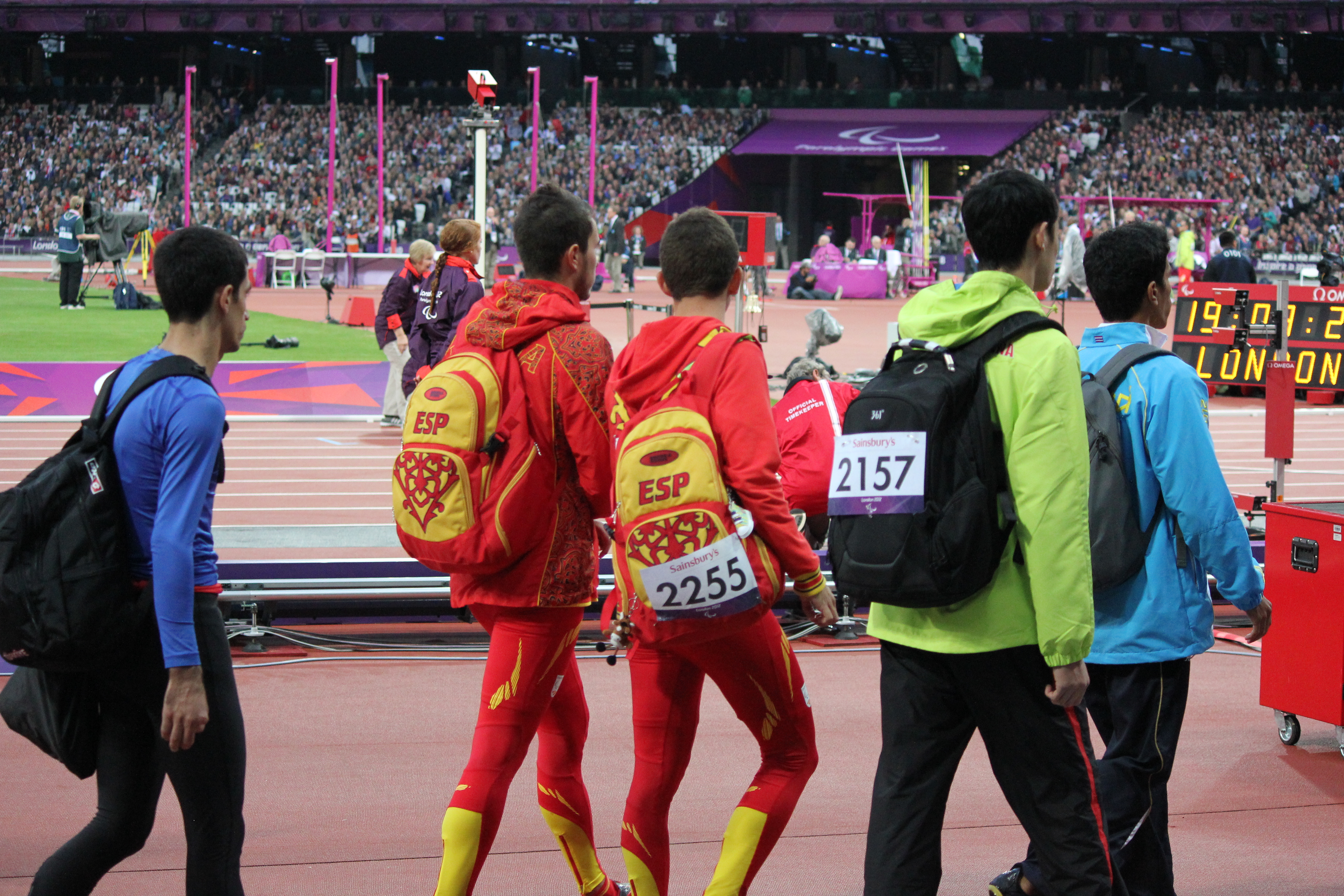}} & \adjustbox{padding=0cm 0.1cm}{\includegraphics[height=1.7cm,valign=c]{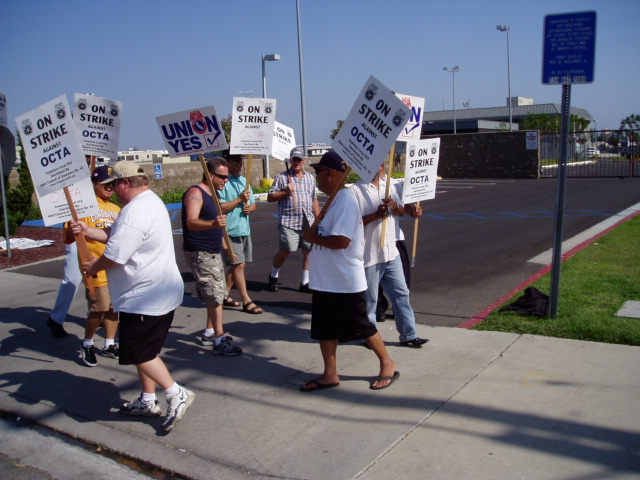}}\\
\midrule
\textbf{Text} & \raggedright\arraybackslash\scriptsize{Bathum and his guide discuss the race following their \underline{\textbf{downhill}} ride.} & \raggedright\arraybackslash\scriptsize{Interior shot of the parliament's \underline{\textbf{Louise Weiss}} building.} &\raggedright\arraybackslash\scriptsize{\underline{\textbf{Spanish}} runners enter the field of play for the start of 400m T12 Men Heat 3.} & \raggedright\arraybackslash\scriptsize{\underline{\textbf{Teamsters}} on strike against OCTA.}\\
\midrule
\textbf{Text-only prediction}&\centering\arraybackslash\scriptsize{\textcolor{red}{Downhill mountain biking}} & \centering\arraybackslash\scriptsize{\textcolor{red}{Louise Weiss}} & \centering\arraybackslash\scriptsize{\textcolor{red}{Spanish language}} & \centering\arraybackslash\scriptsize{\textcolor{red}{International Brotherhood of Teamsters}}\\ 
\midrule
\textbf{\ours prediction} &\centering\arraybackslash\scriptsize{\textcolor{deepgreen}{Downhill (ski competition)}} &\centering\arraybackslash\scriptsize{\textcolor{deepgreen}{Seat of the European Parliament in Strasbourg}} & \centering\arraybackslash\scriptsize{\textcolor{deepgreen}{Spain}}&\centering\arraybackslash\scriptsize{\textcolor{red}{International Brotherhood of Teamsters}}\\
\midrule
\textbf{Ground Truth} & \centering\arraybackslash\scriptsize{Downhill (ski competition)} &\centering\arraybackslash\scriptsize{Seat of the European Parliament in Strasbourg} & \centering\arraybackslash\scriptsize{Spain} &\centering\arraybackslash\scriptsize{Teamsters}\\
\bottomrule 
\end{tabular}
\end{table*}

To delve into the effectiveness of all the modules in our framework, we designed two groups of experiments on three datasets for the ablation study. Table \ref{tab:exp-ablation} presents the results of the experiment.
In the first group, we compared \ours with the following variants: 
(1) w/o Multi-choice Selection: 
when disabling the multi-choice selection module, we simply considered the Top-1 entity from the reranking results in the embedding retrieval as the answer. 
Although embedding retrieval yields impressive results, using multi-choice selection improves the accuracy of \ours on three datasets, particularly on the Wikidiverse dataset, which has complex entity categories. 
This improvement suggests that linking mentions to entities involves more than semantic similarity between descriptions. 
Fine-tuning the LLM enables it to grasp subtler relational logic for effective linking.
(2) w/o Retrieval Augmentation: 
without retrieval augmentation, we applied the fuzzy string matching between entities and mentions to retrieve Top-k candidate entities. 
The significant performance degradation caused by disabling this module emphasizes the vital role of retrieval augmentation. 
Nonetheless, the fact that performance still surpasses all baselines even without this module indirectly highlights the effectiveness of the other modules in \ours.
(3) w/o Entity Augmentation: 
we employed the original entity descriptions instead of their augmented counterparts. 
Augmented entities contribute to improved overall performance, indicating that entity augmentation not only reduces the computational cost of the LLM inference but also yields more precise entity descriptions.
(4) w/o Mention Augmentation: 
Using only basic information of mentions (e.g., names and textual contexts) for entity linking also leads to decreased performance, thereby demonstrating that mention augmentation effectively supplements mentions with valuable information for subsequent successful candidate entities retrieval.
(5) w/o E\&M Augmentation: 
combining (3) and (4), without any augmentations to mentions and entities further impair the performance, with the Top-1 accuracy decreasing by 7.4\%, 3.8\%, and 7.1\% on three datasets, respectively. 
This further confirms the effectiveness of the augmentation modules.

In the second group, we explored the impact of visual modality on our performance. 
(1) Firstly, without using visual information during augmentation, performance declines by 6.2\%, 1.7\%, and 2.8\% on three datasets, respectively. 
This suggests the significance of visual information in entity linking, as it helps mitigate textual context insufficiency-induced ambiguities and enrich descriptive details for mentions.
(2) Furthermore, we assessed the impact of removing visual information on the embedding retrieval performance by disabling the multi-selection module.
Clearly, in the absence of visual information augmentation, the performance of embedding retrieval decreased by 13.3\%, 1.7\%, and 1.5\% on three datasets, respectively.

\subsection{Case Study}

To clearly demonstrate the capabilities of our \ours, we compare the prediction results for textual modality and multimodality in Table \ref{tab:case-study}. 
Cases 1-3 illustrate that predictions based on text-only modality information can be incorrect, whereas multimodal information leads to correct results. 
In Case 1, with text-only modality information, lacking sufficient contextual clues, the model incorrectly links "\textit{downhill}" to "\textit{Downhill mountain biking}", likely influenced by words like "\textit{race}" and "\textit{ride}". 
However, when image information is introduced, \ours captures the "\textit{snowy slope}" depicted in the image and correctly links the mention to "\textit{Downhill (ski competition)}". 

In Case 4, the prediction results are incorrect in both the text-only and multimodal modalities. 
This could be due to the model overinterpreting both text and image information, leading to an excessively detailed selection of linked entities.

\section{CONCLUSION}
In this paper, we proposed a unified framework (\ours) which establishes a new paradigm to process MEL tasks using LLMs.
Concretely, we leveraged the capabilities of LLMs to augment the representation of mentions and entities individually by integrating textual and visual information and refining textual information.
Then, we employed the embedding-based method to narrow down the candidate set. 
Afterwards, we drove the LLM as a selector to choose the entity from the candidate set that best matches the mention.
Extensive experiments on three public datasets have validated
the effectiveness of our \ours framework compared with several
state-of-the-art baseline methods.




\bibliographystyle{ACM-Reference-Format}
\bibliography{reference}


\end{document}